\documentclass[letterpaper]{article}

\usepackage{iccc}
\usepackage[utf8]{inputenc} 
\usepackage{times}          
\usepackage{helvet}         
\usepackage{courier}        
\usepackage{CJKutf8}        
\usepackage{url}
\usepackage{hyperref}
\usepackage[table]{xcolor}  
\usepackage{graphicx}       
\usepackage{caption}        
\usepackage{subcaption}     
\usepackage{booktabs}       
\usepackage{multicol}       
\usepackage{makecell}       
\usepackage{arydshln}       
\usepackage{comment}        
\usepackage{stfloats}       
\usepackage{adjustbox}      

\newcommand{\ukiyoe}{Ukiyo-e}
\newcommand{\ukiyoeJP}{浮世絵}
\newcommand{\ARCFullName}{ARC Ukiyo-e Collection}

\newcommand{\UkiyoeFDFullName}{Ukiyo-e Faces Dataset}
\newcommand{\Kaokore}{Kaokore}
\newcommand{\KaokoreFullNameEn}{Collection of Facial Expressions}
\newcommand{\Rekognition}{Amazon Rekognition}
\newcommand{\pixtopixhd}{Pix2PixHD}
\newcommand{\superth}{\textsuperscript{th}}

\newcommand{\todo}[1]{}
\renewcommand{\todo}[1]{{\color{red} TODO: {#1}}}  

\definecolor{hirosada}{rgb}{0.00, 0.11, 0.50}
\definecolor{kogyo}{rgb}{0.69, 0.25, 0.05}
\definecolor{kunichika}{rgb}{0.07, 0.44, 0.11}
\definecolor{kunisada1stgen}{rgb}{0.55, 0.03, 0.00}
\definecolor{kunisada2ndgen}{rgb}{0.35, 0.12, 0.44}
\definecolor{kunisada3rdgen}{rgb}{0.35, 0.18, 0.05}
\definecolor{kuniyoshi}{rgb}{0.64, 0.21, 0.51}
\definecolor{toyokuni1stgen}{rgb}{0.24, 0.24, 0.24}
\definecolor{toyokuni3rdgen}{rgb}{0.72, 0.52, 0.04}
\definecolor{yoshitaki}{rgb}{0.00, 0.39, 0.45}

\definecolor{c_photo}{rgb}{0.09, 0.33, 0.30}
\definecolor{c_ukiyo_e}{rgb}{0.63, 0.47, 0.29}
\definecolor{c_kaokore}{rgb}{0.78, 0.71, 0.93}

\definecolor{lm_choose}{rgb}{0.56, 1.00, 0.56}
\definecolor{lm_reject}{rgb}{1.00, 0.74, 0.61}
\definecolor{lm_special}{rgb}{0.73, 0.67, 0.99}

\definecolor{link}{rgb}{0.02,0.27,0.67}

\graphicspath{ {./} }

\newcommand{\ks}{\vspace*{-5.2px}} 
\newcommand{\citenb}[1]{ \citeauthor{#1}~\shortcite{#1} } 

\title{Ukiyo-e Analysis and Creativity with Attribute and Geometry Annotation}
\author{Yingtao Tian\\
Google Brain\\
Tokyo, Japan
\And
Tarin Clanuwat \\
ROIS-DS Center for \\Open Data in the Humanities\\
NII
\And
Chikahiko Suzuki \\
ROIS-DS Center for \\Open Data in the Humanities\\
NII
\And
Asanobu Kitamoto \\
ROIS-DS Center for \\Open Data in the Humanities\\
NII
}

\begin{document} 
\begin{CJK}{UTF8}{min}

\maketitle

\begin{abstract}
    The study of \ukiyoe{}, an important genre of pre-modern Japanese art, focuses on the \emph{object} and \emph{style} like other artwork researches.
    Such study has benefited from the renewed interest by the machine learning community in culturally important topics, leading to interdisciplinary works including collections of images, quantitative approaches, and machine learning-based creativities.
    They, however, have several drawbacks, and it remains challenging to integrate these works into a comprehensive view.
    To bridge this gap, we propose a holistic approach\footnote{\url{https://github.com/rois-codh/arc-ukiyoe-faces}}:
    We first present a large-scale \ukiyoe{} dataset with coherent semantic labels and geometric annotations, then show its value in a quantitative study of \ukiyoe{} paintings' \emph{object} using these labels and annotations. 
    We further demonstrate the machine learning methods could help \emph{style} study through soft color decomposition of \ukiyoe{},
    and finally provides joint insights into \emph{object} and \emph{style} by composing sketches and colors using colorization.
\end{abstract}

\section{Introduction}

\begin{figure}[ht]
\centering

\ks{}

\begin{minipage}{0.38\columnwidth}
    \centering
    \includegraphics[trim=0 75px 0 25px, clip,width=\textwidth]{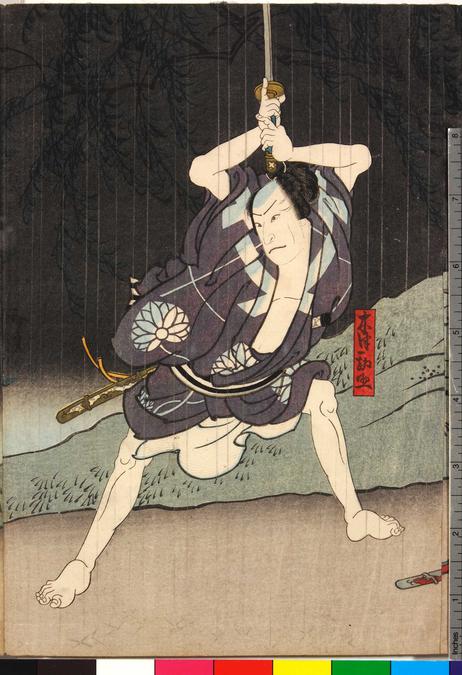}
\end{minipage}
\hfill
\begin{minipage}{0.60\columnwidth}
    \centering
        \begin{small}
        \begin{tabular}{cc}
        \toprule
        Attribute & Value \\
        \midrule
        \makecell{Title\\(画題)}  &  \makecell{\emph{Kizukansuke}\\(「木津勘助」)} \\
        \hdashline
        \makecell{Painter\\(絵師)}  & \textcolor{hirosada}{\makecell{Hirosada\\(広貞)}} \\
        \hdashline
        \makecell{Format\\(判種)}  & \makecell{Middle-size / \emph{Nishiki-e}\\ (中判/錦絵)} \\
        \hdashline
        \makecell{Year in AD\\(西暦)}  & 1849 \\
        \hdashline
        \multicolumn{2}{c}{$\cdots\cdots$} \\
        \bottomrule
        \end{tabular}
        \end{small}
\end{minipage}

\ks{}

\caption{An example of \ukiyoe{} work in \ARCFullName{} (Object Number \href{https://www.dh-jac.net/db/nishikie/results-big.php?f1[]=arcUP2451}{arcUP2451} ) titled \emph{Kizukansuke} by painter \textcolor{hirosada}{Hirosada}.
The painting on the left is accompanied by metadata for this work on the right.
For example, metadata further indicates this work is a middle-sized \emph{Nishiki-e} (multi-colored woodblock printing) produced in 1849.}
\label{fig:arc.exmaple}

\ks{}

\vspace*{-2px}

\end{figure}

The Edo period of Japan (16\superth{} to 19\superth{} century) has seen the prosper of \ukiyoe{} (\ukiyoeJP{}), a genre of pre-modern Japanese artwork that consists of paintings and woodblock printings.
Unlike early dominating Emakimono (絵巻物, \emph{picture scroll}) and Ehon (絵本, \emph{picture book}) that focus on famous figures and stories in Sinosphere culture and classic Japanese stories, the topic of \ukiyoe{} extends broadly to daily subjects, 
such as characters like beauties and \emph{Kabuki} (歌舞伎), landscape arts, animals and plants in everyday life, and even contemporary news.
As an example, Figure~\ref{fig:arc.exmaple} shows a \ukiyoe{} depicting a \emph{Kabuki} performance.
The popularity of woodblock printing makes it possible to produce paintings on a larger scale at a lower cost, which contributes to the flourish of \ukiyoe{} and leaves us with a vast collection of artworks in this genre~\cite{kobayashi1994ukiyoe,ius2008encyclopedia}.
Such an extensive and varied collection provides a valuable corpus for Japanese artwork research.

The subject of such artwork study could be multi-faceted involving several aspects, of which two crucial are the \emph{object} in the painting, such as the outline and the shape of depicted figures, and the \emph{style} of painting, such as textures and colors.
For example, the former reveals the trend of objects depicted over time, and the latter allows the identification of artists~\cite{chikahiko18collection}.
The renewed interest by the machine learning community in the culturally essential topics has led to works addressing the traditional Japanese artworks from an interdisciplinary perspective.
Along this line of research, building open collections of digitized images has been proposed for Ehon~\cite{chikahiko18collection} and \ukiyoe{}~\cite{arc2005,pinkney2020ukiyoe}.
Further works use quantitative approaches into the \emph{object} for artworks, such as studying the geometry features of Buddha statues~\cite{renoust2019} and \ukiyoe{} faces~\cite{renoust2019},
Alternatively, inspired by the art nature of painting, machine learning-based creativity has been leveraged for studying \emph{style}, such as painting process generation~\cite{tian2020kaokore} and image synthesis across artwork and photorealistic domains~\cite{pinkney2020resolution}.
These works provide valuable connections between machine learning and the humanities research of 
Japanese artwork.

We, however, also notice that these works present several drawbacks. 
For example, collection on digitized images may either comes with no semantic~\cite{pinkney2020ukiyoe} or is in a format not designed with machine learning-based applications in mind.
Furthermore, quantitative approaches are only conducted on a small set of artworks~\cite{murakami2007} or require extensive human labor to adapt for \ukiyoe{}~\cite{renoust2019}, and machine learning-based creativity works may deal more with cross-domain art expression~\cite{pinkney2020resolution} than the very domain of artwork on which humanities research focuses.
Finally, the art study into a particular genre requires insights into both the \emph{object} and \emph{style} to acquire a comprehensive understanding.
Current works, however, only address one of the \emph{object} or \emph{style}, falling short of the expectation. 

To overcome the aforementioned drawbacks and to provide deeper insight into the artistic style of \ukiyoe{},
we propose a new approach that is (1) holistic in both studying the \emph{object} and \emph{style} through the joint use of images, labels, and annotations, and (2) powered by large scale data and state-of-the-art machine learning model than the prior works.
To summarize, our main contributions are as follow:
\begin{itemize}
    \item We present a large-scale ($11,000$ paintings and $23,000$ faces) \ukiyoe{} dataset with coherent semantic labels and geometric annotations, through augmenting and organizing existing datasets with automatical detection.
    \item We are the first to conduct a large-scale quantitative study of \ukiyoe{} paintings (on more than $11,000$ paintings), providing understanding into \emph{object} in artworks by jointly quantifying semantic labels and geometric annotations.
    \item We show that machine learning-based models could provide insights into \emph{style} by decomposing finished \ukiyoe{} images into color-split woodblocks that reflect how \ukiyoe{} images were possibly produced.
    \item We study and show machine learning-based creativity model could engage problems that arise jointly studying \emph{object} and \emph{style} by separating geometry shapes and artistic styles in an orthogonal and re-assemblable way.
\end{itemize}

\section{Dataset}

Art research in traditional paintings often asks questions regarding the work, like the author and production year.
One focus in such research is on faces since they could help answer these questions through quantitative analysis.
In this direction, \KaokoreFullNameEn{}~\cite{chikahiko18collection,tian2020kaokore} provides a large-scale ($8848$ images) set of coarse-grained cropped faces.
Another study~\cite{murakami2007} deals with facial landmarks which are more fine-grained than cropped faces to support quantitative analysis.
However, its manual labeling process only allows analysis on a small set (around $50$ images) of \ukiyoe{} paintings.

\begin{figure}[ht!]
\centering

\ks{}

\hfill
\begin{subfigure}[b]{0.3\columnwidth}
    \centering
    \includegraphics[width=\columnwidth]{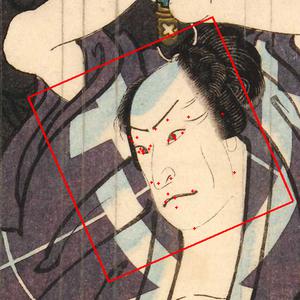}
\end{subfigure}\hfill    
\begin{subfigure}[b]{0.3\columnwidth}
    \centering
    \includegraphics[width=1\columnwidth]{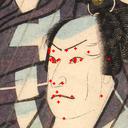}
\end{subfigure}\hfill    
\hfill

\vspace*{5px}

\begin{subfigure}[b]{1.0\columnwidth}
    \centering
    \begin{adjustbox}{width=1.0\textwidth}
        \begin{tabular}{ll}
        \toprule
            Facial Region & Landmarks \\
        \midrule
            Left Eye      & Center, Left, Right, Up, Down \\
            Right Eye     & Center, Left, Right, Up, Down \\
            Left Eyebrow  & Left, Right, Up \\
            Right Eyebrow & Left, Right, Up \\
            Left Pupil    & Center \\
            Right Pupil   & Center \\
            Mouth         & Left, Right, Up, Down \\
            Nose          & Center, Left, Right \\
            Jawline       & Upper Left \& Right, Mid Left \& Right, Chin Bottom \\
        \bottomrule
        \end{tabular}
    \end{adjustbox}
\end{subfigure}\hfill        

\ks{}

\caption{An exmaple of detected landmarks and the extracted face in Figure~\ref{fig:arc.exmaple}'s \ukiyoe{} painting.
On the left, the red dots show detected facial landmarks and the rectangle shows the bounding box inferred from these landmarks. 
The right image shows the extracted face from the bounding box.
The table lists a summary of all landmark locations. 
}
\label{fig:arc.example.annoated}

\ks{}

\end{figure}
\begin{figure}[ht!]

\centering



\begin{subfigure}[t]{0.9\columnwidth}
    \includegraphics[trim=0 0 0 128, clip, width=\textwidth]{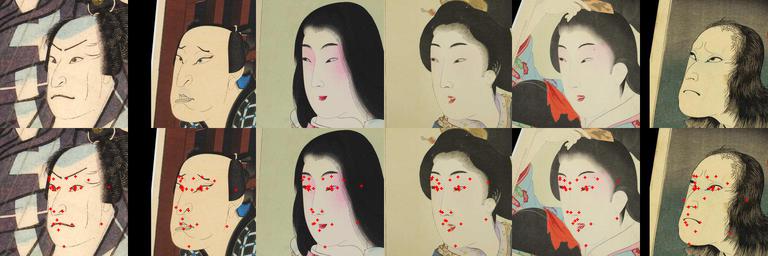}
\end{subfigure}

\vspace{3px}

\begin{subfigure}[t]{0.9\columnwidth}
    \includegraphics[trim=0 0 0 128, clip, width=\textwidth]{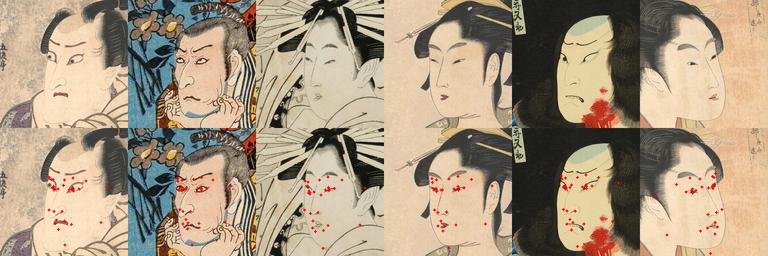}
\end{subfigure}

\ks{}

\caption{Faces with their landmarks. 
In each row, we show six examples of extracted faces annoated with their corresponding landmarks in the same format as Figure~\ref{fig:arc.example.annoated}.
}

\label{fig:extracted.face.grid}

\ks{}

\end{figure}
\begin{figure*}[bt!]
\centering

\ks{}

\begin{subfigure}[t]{0.56\textwidth}
    \includegraphics[width=\textwidth]{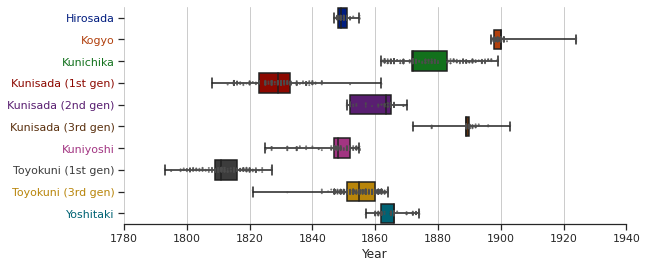}
    \caption{Distribution of years with respect to authors.}
\end{subfigure}\hfill
\begin{subfigure}[t]{0.40\textwidth}
    \includegraphics[width=\textwidth]{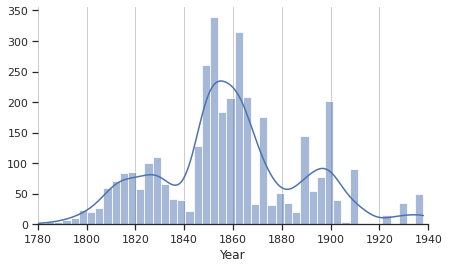}
    \caption{Distribution of years of all works in the dataset.}
\end{subfigure}

\newcommand{\imgwidth}{5.5in}
\begin{subfigure}[t]{0.95\textwidth}
    \begin{center}
        \begin{small}
            \begin{tabular}{rp{\imgwidth}}
                \toprule
                \textit{Painter}   & \textit{Examples} \\     

                \midrule
                \textcolor{hirosada}{\makecell[r]{Hirosada\\(広貞)}} & \parbox[c]{2em}{
                    \includegraphics[width=\imgwidth]{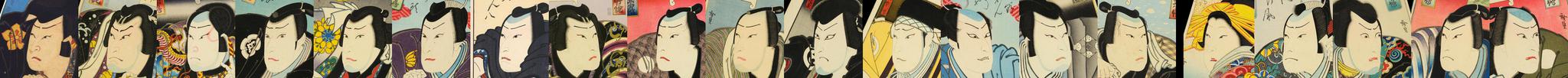}} \\

                \midrule
                \textcolor{kogyo}{\makecell[r]{Kogyo\\(耕漁)}} & \parbox[c]{2em}{
                    \includegraphics[width=\imgwidth]{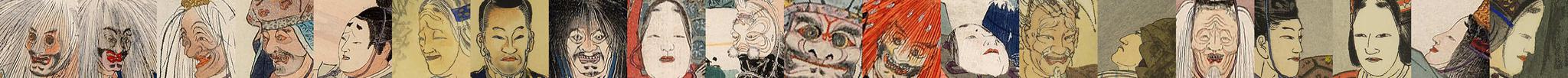}} \\

                \midrule
                \textcolor{kunichika}{\makecell[r]{Kunichika\\(国周)}} & \parbox[c]{2em}{
                    \includegraphics[width=\imgwidth]{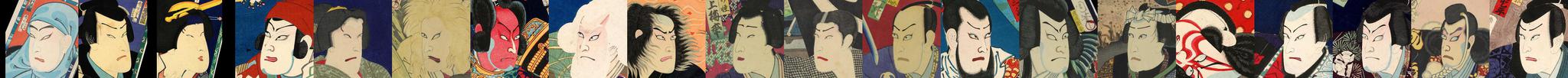}} \\

                \midrule
                \textcolor{kunisada1stgen}{\makecell[r]{Kunisada (1st gen)\\(国貞 初代)}} & \parbox[c]{2em}{
                    \includegraphics[width=\imgwidth]{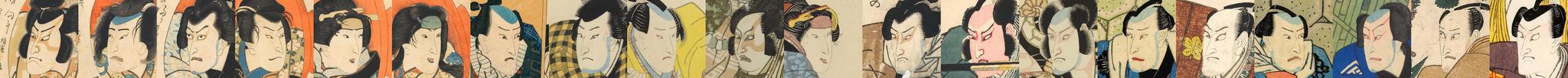}} \\

                \midrule
                \textcolor{kunisada2ndgen}{\makecell[r]{Kunisada (2nd gen)\\(国貞 二代目)}} & \parbox[c]{2em}{
                    \includegraphics[width=\imgwidth]{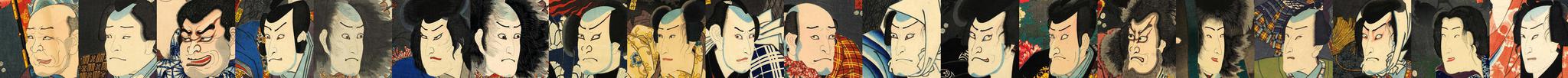}} \\

                \midrule
                \textcolor{kunisada3rdgen}{\makecell[r]{Kunisada (3rd gen)\\(国貞 三代目)}} & \parbox[c]{2em}{
                    \includegraphics[width=\imgwidth]{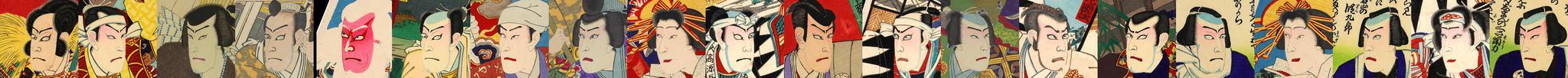}} \\

                \midrule
                \textcolor{kuniyoshi}{\makecell[r]{Kuniyoshi\\(国芳)}} & \parbox[c]{2em}{
                    \includegraphics[width=\imgwidth]{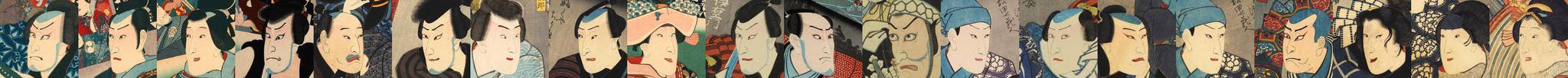}} \\

                \midrule
                \textcolor{toyokuni1stgen}{\makecell[r]{Toyokuni (1st gen)\\(豊国 初代)}} & \parbox[c]{2em}{
                    \includegraphics[width=\imgwidth]{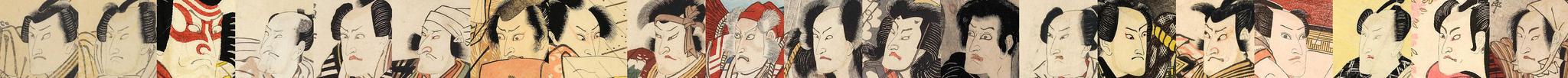}} \\

                \midrule
                \textcolor{toyokuni3rdgen}{\makecell[r]{Toyokuni (3rd gen)\\(豊国 三代目)}} & \parbox[c]{2em}{
                    \includegraphics[width=\imgwidth]{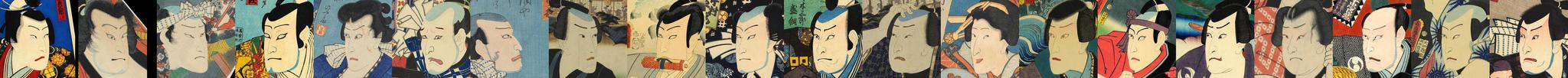}} \\
                    
                \midrule
                \textcolor{yoshitaki}{\makecell[r]{Yoshitaki\\(芳滝)}} & \parbox[c]{2em}{
                    \includegraphics[width=\imgwidth]{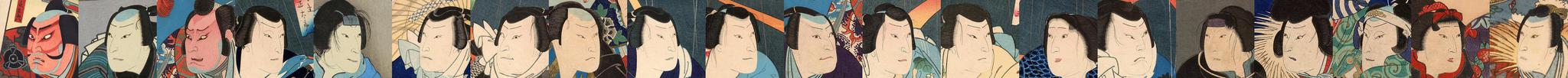}} \\
                
                \bottomrule
            \end{tabular}
        \end{small}
    \end{center}
    \caption{Example of paintings, represented by the extracted faces, by authors.}
\end{subfigure}

\ks{}

\caption{Metadata and their associated paintings.
We jointly show two important metadata, year and author.
We also show show ten authors with the most painting in the dataset.
\textbf{(a)} illustrates the year distribution with respect to authors, and \textbf{(b)} shows the overall year distribution. 
Exemplary paintings belonging to these authors are shown in \textbf{(c)}.}

\label{fig:year.distribution}

\ks{}

\end{figure*}

To combine both works' advantage, we extend existing datasets through augmentation and automated annotation, resulting in a large-scale \ukiyoe{} dataset with a more fine-grained facial feature. 
The rest of this section details the process and analysis of our new proposed dataset.

\subsection{Fundamental Datasets}

We build our work based on two foundation datasets.
One of them is \ARCFullName{}~\cite{arc2005}, a publicly available service that provides access to digitized \ukiyoe{} paintings primarily in the Edo period, plus metadata compiled by domain experts.
It has $11,103$ entries of painting and the associated metadata, one example of which is shown in Figure~\ref{fig:arc.exmaple}.
This service allows researchers to dive into curated metadata for comparative study for art research.

Another dataset is \UkiyoeFDFullName{}~\cite{pinkney2020ukiyoe}, a public available dataset of \ukiyoe{} faces extracted from \ukiyoe{} images available online.
With $5,000$ high-quality faces, this dataset plays an essential role in controllable image generation across \ukiyoe{} faces and photo-realistic human faces~\cite{pinkney2020resolution}.
However, as this dataset focuses on image synthesis, it does not include metadata for \ukiyoe{} paintings from which faces are extracted.

\subsection{Geometric Annotation with Facial Landmark Detection}

Inspired by \citenb{pinkney2020ukiyoe}, we use an face recognition API, \Rekognition{} (\href{https://aws.amazon.com/rekognition/}{\color{link} link}), to detect facial landmarks in in \UkiyoeFDFullName{} paintings. 
Despite targeting photo-realistic human face images, this API demonstrates compelling accuracy on \ukiyoe{} paintings.
Since the detected faces may not be well-aligned, we infer the possibly rotated bounding box of faces for cropping faces from the painting, inspired by the preprocessing in FFHQ~\cite{karras2019style}.
In Figure~\ref{fig:arc.example.annoated} we show an example of detected landmarks and the face extraction process.

A total of $18,921$ faces and their corresponding facial landmarks have been detected from paintings in \ARCFullName{}.
Furthermore, since \UkiyoeFDFullName{}~\cite{pinkney2020ukiyoe} also follows the same preprocessing as FFHQ, its $5,000$ faces are comparable to the faces extracted from \ARCFullName{}.
Although faces in \UkiyoeFDFullName{} lack metadata,  we can still incorporate them for geometry statistics by going through the above-mentioned landmark detecting process.
In doing so, we have a total of around $23,000$ \ukiyoe{} faces.
In Figure~\ref{fig:extracted.face.grid}, we show examples of such faces and their landmarks.  

\subsection{Semantic Labels Incorporation}

\begin{figure}[!ht]
\centering

\ks{}

\begin{subfigure}[t]{0.30\columnwidth}
    \includegraphics[width=\textwidth]{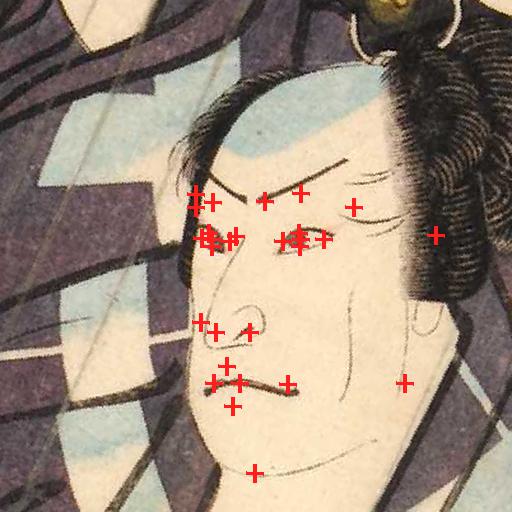}
    \caption{Landmarks de\-tected auto\-matically }
\end{subfigure}\hfill
\begin{subfigure}[t]{0.30\columnwidth}
    \includegraphics[width=\textwidth]{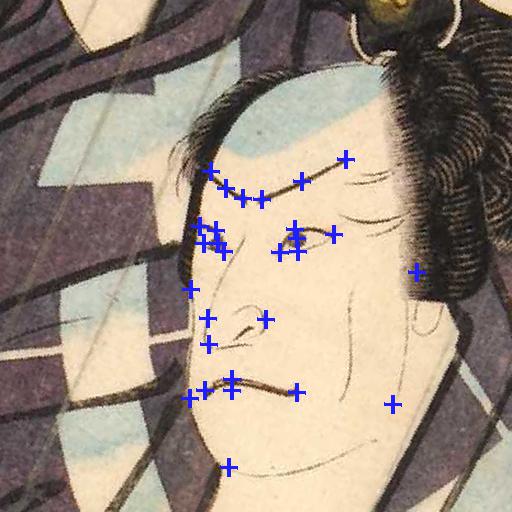}
    \caption{Landmakrs ma\-nually annotated by domain experts}
\end{subfigure}\hfill
\begin{subfigure}[t]{0.30\columnwidth}
    \includegraphics[width=\textwidth]{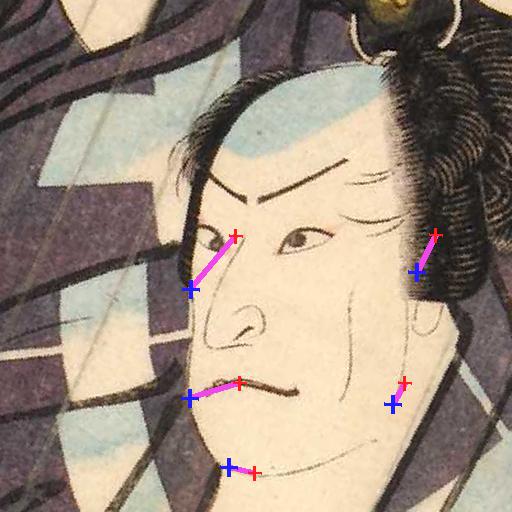}
    \caption{Detailed Study in Jawline landmarks}
\end{subfigure}

\begin{subfigure}[b]{1.0\columnwidth}
    \centering
    \begin{adjustbox}{width=0.95\textwidth}
        \begin{tabular}{ll}
            \toprule
            Facial Region & Landmarks (Mean Error in Pixel Distance) \\   
            \midrule  
            Left Eye       & \colorbox{lm_choose}{Center ($10.2$)} \\
            Right Eye      & \colorbox{lm_choose}{Center ($18.5$)} \\                
            Mouth          & \colorbox{lm_choose}{Left ($9.7$)} \colorbox{lm_choose}{Right ($13.4$)} \\
            Nose           & \colorbox{lm_reject}{Center ($22.1$)} \colorbox{lm_choose}{Left ($18.6$)} \colorbox{lm_choose}{Right ($16.5$)} \\
            Left EyeBrow   & \colorbox{lm_reject}{Left ($34.2$)} \colorbox{lm_reject}{Right ($43.0$)} \colorbox{lm_reject}{Up ($23.1$)} \\
            Right EyeBrow  & \colorbox{lm_choose}{Left ($17.4$)} \colorbox{lm_reject}{Right ($41.8$)} \colorbox{lm_reject}{Up ($53.1$)} \\
            Jawline *       & \colorbox{lm_reject}{Upper Left ($69.4$)} \colorbox{lm_choose}{Upper Right ($57.4$)} \colorbox{lm_reject}{Mid Left ($25.0$)} \\
                            & \colorbox{lm_choose}{Mid Right ($56.2$)} \colorbox{lm_choose}{Chin Bottom ($57.8$)} \\
            \bottomrule
        \end{tabular}
    \end{adjustbox}
    \caption{Landmarks with the mean error in pixel distance between detected and expert labeled postion. \colorbox{lm_choose}{Landmarks in green} are considered of high-quality  and \colorbox{lm_reject}{those in red} of low-quality.}
\end{subfigure}\hfill


\caption{Study of landmark quality by comparing automatically detected positions \textbf{(a)} with expert labeled positions \textbf{(b)}.
As \ukiyoe{} faces are mostly towards either left or right, we normalize all paintings to face left for the geometry purpose.
The study has been conducted for $69$ \ukiyoe{} paintings, and the mean error of pixel distances are aggregated in \textbf{(d)}, which we use to decide which landmarks are considered high-quality. 
The decision is based on picking landmarks with a low error of pixel distance (heuristically those $<20$), except for Jawline (*) that needs special consideration: as \textbf{(c)} shows, 
landmarks on the the direction of facing (Upper Left, Mid Left) are useless since they are invisible in most \ukiyoe{}. 
The others (Upper Right, Mid Right, Chin Bottom) are valuable since they still lie on the jawline, and they are far from other landmarks, allowing larger error marging when used for calculating angular features.
}
\label{fig:userstudy}

\ks{}

\end{figure}
\begin{figure}[ht]
\centering

\ks{}

\begin{subfigure}[t]{0.45\columnwidth}
    \centering
    \includegraphics[width=0.7\textwidth]{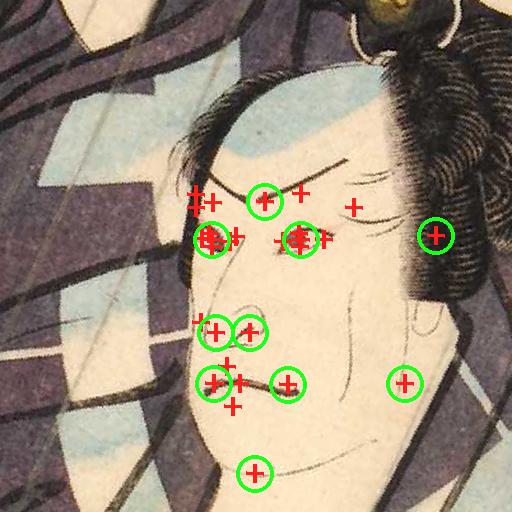}
    \caption{High quality landmarks circled in green.}
\end{subfigure}\hfill
\begin{subfigure}[t]{0.5\columnwidth}
    \centering
    \includegraphics[width=0.63\textwidth]{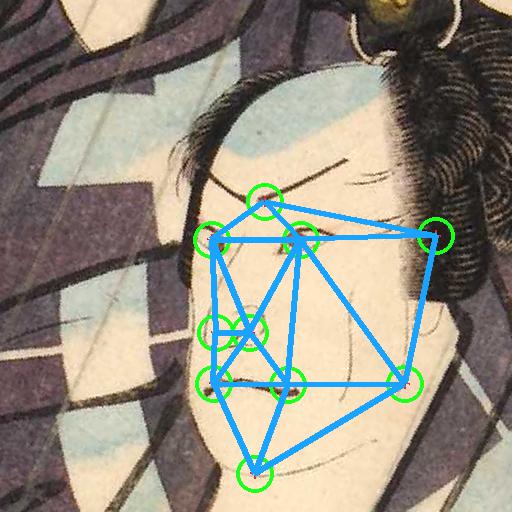}
    \caption{Examples of angles formed by high quality landmarks}
\end{subfigure}

\ks{}

\caption{Extracted geometry features from high quality landmarks.
In (a) we highlight landmarks of high quality,
and in (b) we show some a subset of of angles formed by high quality landmarks for brevity.
}
\label{fig:arc.qaunt}

\ks{}

\end{figure}

As our dataset is derived from \ARCFullName{},
we can also relate faces and the corresponding landmarks with the original metadata, 
such as the year of creation and the author of the painting.
In Figure~\ref{fig:year.distribution} we show these two metadata jointly, as well as exemplary paintings belonging to several authors.
For example, we can observe Shumei (襲名, \emph{name succession}) system common in traditional Japanese art community where an artist takes his/her teacher's name, as the case of the lineage of \textcolor{kunisada1stgen}{Kunisada 1st gen (国貞 初代)}, \textcolor{kunisada2ndgen}{Kunisada 2nd gen (国貞 二代目)} and \textcolor{kunisada3rdgen}{Kunisada 3rd gen (国貞 三代目)}.
Furthermore, we can also notice three peaks of production of \ukiyoe{} painting, occupying the early, mid, and late 19\superth{} century.
The last peak is dominated by \textcolor{kogyo}{{Kogyo (耕漁)}} who painted well into the 20\superth{} century and whose uniqueness is further shown in his exemplary paintings under the influence of modern painting.
\begin{figure*}[!t]
\centering

\ks{}

\newcommand{\subcaptionspace}{\vspace{-20px}}
\newcommand{\subwidth}{0.21\textwidth}

\begin{subfigure}[b]{\subwidth{}}
    \includegraphics[width=\textwidth]{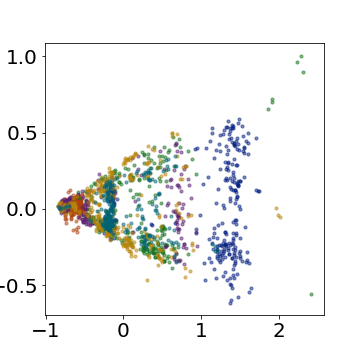}
    \subcaptionspace{}
    \subcaption*{PCA}
\end{subfigure}\hfill
\begin{subfigure}[b]{\subwidth{}}
    \includegraphics[width=\textwidth]{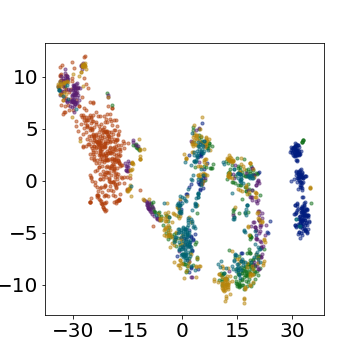}
    \subcaptionspace{}
    \subcaption*{T-SNE}
\end{subfigure}\hfill
\begin{subfigure}[b]{\subwidth{}}
    \includegraphics[width=\textwidth]{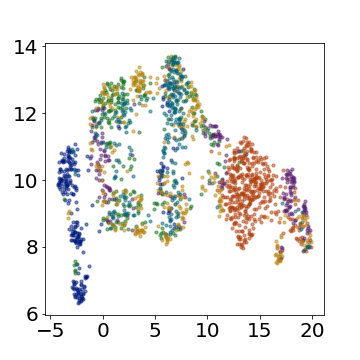}
    \subcaptionspace{}
    \subcaption*{UMAP}
\end{subfigure}\hfill
\begin{subfigure}[b]{\subwidth{}}
    \includegraphics[width=\textwidth]{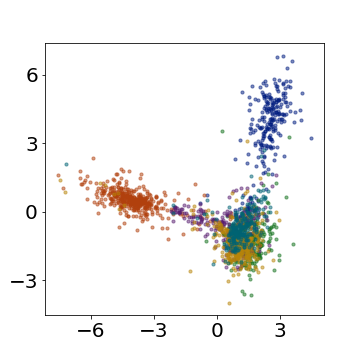}
    \subcaptionspace{}
    \subcaption*{LDA}
\end{subfigure}\hfill
\begin{subfigure}[b]{0.15\textwidth}  
    \includegraphics[trim=2px 2px 2px 2px, clip=true, width=\textwidth]{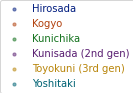}
    \vspace{5px}
    \subcaption*{(Legend)}
\end{subfigure}\hfill

\ks{}

\caption{Unsupervised (PCA, T-SNE, UMAP) and supervised (LDA) clustering of faces' geometry features in a two-dimensional plane. 
We show works by six most-frequently appearing authors in the clustering.
Labels are used for coloring the authors for illustrative purposes only, and are not used in the clustering except for LDA.
Visually, \textcolor{hirosada}{Hirosada} and \textcolor{kogyo}{Kogyo} are shown with clear separation from other authors, which could be cross-verified with explanation using other information. 
}
\label{fig:arc.plot.author}

\ks{}

\end{figure*}
\begin{figure}[ht]
\centering

\ks{}

\newcommand{\subcaptionspace}{\vspace{-15px}}
\newcommand{\subwidth}{0.32\columnwidth}

\begin{subfigure}[t]{\subwidth}
    \includegraphics[width=\textwidth]{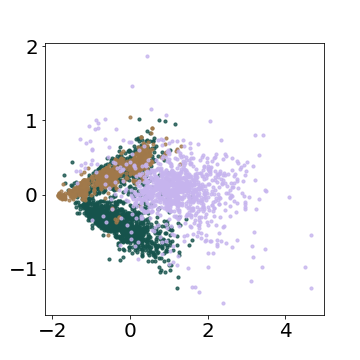}
    \subcaptionspace{}
    \subcaption*{PCA}
\end{subfigure}\hfill
\begin{subfigure}[t]{\subwidth}
    \includegraphics[width=\textwidth]{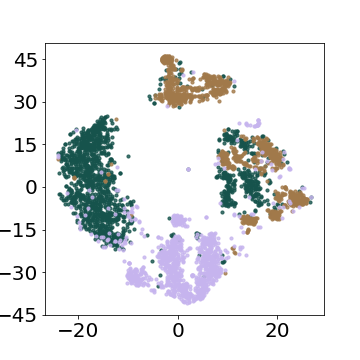}
    \subcaptionspace{}
    \subcaption*{T-SNE}
\end{subfigure}\hfill
\begin{subfigure}[t]{\subwidth}
    \includegraphics[width=\textwidth]{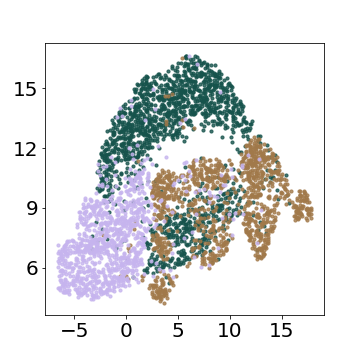}
    \subcaptionspace{}
    \subcaption*{UMAP}
\end{subfigure}\hfill

\ks{}

\caption{Unsupervised (PCA, T-SNE, UMAP) clustering of faces using geometry features for \textcolor{c_ukiyo_e}{\ukiyoe{} paintings}, \textcolor{c_kaokore}{\Kaokore{} paintings} and \textcolor{c_photo}{photo-realistic human faces}.
Visually, we see a separation between faces of different sources.
}
\label{fig:ukiyoe_vs_kaokore_vs_human.quant}

\ks{}

\end{figure}

\section{Experiment}

\subsection{Study \ukiyoe{} \emph{Object} using Geometry Features}

Regarding the content, art researches may be divided into two categories: 
the shape that deals with geometry features and the texture that deals with brushes and color features. 
To quantitatively provide insights on attributes such as the author and the painting year, either category can be used for unsupervised learning, like clustering, or supervised learning, like predicting metadata,
While the texture features could help analyze attributes for a single work~\cite{tian2020kaokore}, the geometry features could also be considered since the texture may vary due to \ukiyoe{}'s frequent reprint~\cite{murakami2007} or sculpture's preservation condition~\cite{renoust2019},
Both works propose to leverage facial landmarks to infer geometry features such as angles and iconometric proportions to quantify artwork.

However, since both works rely on manually labeled landmarks, they either suffer from being too small (only around $50$ \ukiyoe{} paintings are annotated with landmarks) or require extensive human effort if we ever want to apply the technique used on sculpture to \ukiyoe{}.
To bridge this gap, we propose to use automatically detected landmarks as geometry features. 
To our best knowledge, we are the first to conduct large-scale (more than 10k paintings) quantitative analysis of \ukiyoe{} painting. 
We hope it could serve as an invitation for further quantitative study in artworks.

\subsubsection{Geometry Features from Landmarks}

Inspired by \citenb{murakami2007}, we consider the angles formed by landmarks as they are geometry-invariant under rotation. 
To attain a clear understanding of the quality of landmarks,
we conduct a study on $69$ \ukiyoe{} paintings, comparing landmarks that are automatically detected with positions manually annotated by domain experts, as detailed in Figure~\ref{fig:userstudy}.
We observe that, despite the general high-quality of the predicted landmarks on \ukiyoe{} painting, some landmarks have systematically worse quality than others we decided not to consider.
In the end, we calculate $252$ angles formed by all possible triplets of high-quality landmarks as geometry features for each face, as illustrated in Figure~\ref{fig:arc.qaunt}.

\subsubsection{Analysis on Authorship}
To illustrate the information of geometry features,
we conduct unsupervised (PCA, T-SNE, UMAP) and supervised (LDA) clustering of faces using geometry features in Figure~\ref{fig:arc.plot.author}.
All clusterings show two distinctive authors, \textcolor{kogyo}{Kogyo} and \textcolor{hirosada}{Hirosada (広貞)}, are separated from other authors.
Such separation could be supported through visual inspection into original paintings.
For example, Figure~\ref{fig:year.distribution} (c) shows  \textcolor{kogyo}{Kogyo} and \textcolor{hirosada}{Hirosada} has visually distinctive styles compared to other painters.
Furthermore, such separation could also be cross-verified with analysis leveraging other information sources, 
Figure~\ref{fig:year.distribution} (a) shows that \textcolor{kogyo}{Kogyo} was active well into the 20\superth{} century where in contrast \ukiyoe{} paintings are mainly around the middle 19\superth{} century, 
Furthermore, his uniqueness of style is visually illustrated in exemplary paintings.
We can also observe \textcolor{hirosada}{Hirosada} forms a unique style related to the geographical factors.
While most of the painters analyzed at that time worked in Edo (modern Tokyo), \textcolor{hirosada}{Hirosada} was active in Osaka.
Comparison of culture at that time could be made between the around Edo/Tokyo region, the \emph{de facto} capital of Japan under Tokugawa shogunate, and Kamigata (上方) region encompassing Kyoto and Osaka, the \emph{de jure} capital of Japan and the cultural center of western Japan.
The style of \textcolor{hirosada}{Hirosada} and \textcolor{yoshitaki}{Yoshitaki (芳滝)} who were active in western Japan is therefore called Kamigatae (上方絵, \emph{Kamigata painting}) and is a subject for comparative study. 

\begin{figure*}[!t]
\centering

\ks{}


\newcommand{\signature}{z_arc256-0}
\begin{subfigure}[t]{0.102\textwidth}
    \includegraphics[width=\textwidth]{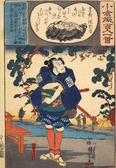}
\end{subfigure}\hfill
\begin{subfigure}[t]{0.715\textwidth}
    \includegraphics[width=\textwidth]{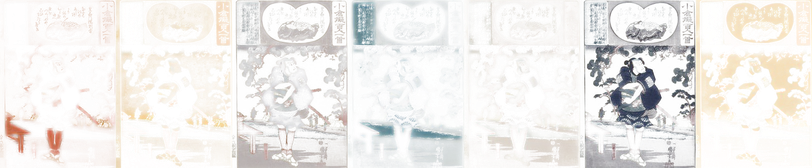}
\end{subfigure}\hfill
\begin{subfigure}[t]{0.102\textwidth}
    \includegraphics[width=\textwidth]{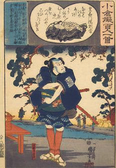}
\end{subfigure}\hfill
\\

\begin{subfigure}[t]{0.102\textwidth}
    \includegraphics[width=\textwidth]{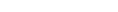}
\end{subfigure}\hfill
\begin{subfigure}[t]{0.715\textwidth}
    \includegraphics[trim=0 0 0 30px, clip, width=\textwidth]{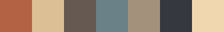}
\end{subfigure}\hfill
\begin{subfigure}[t]{0.102\textwidth}
    \includegraphics[width=\textwidth]{./images/null.png}
\end{subfigure}\hfill

\vspace*{12px}

\renewcommand{\signature}{z_arcface256-1}

\begin{subfigure}[t]{0.102\textwidth}
    \includegraphics[width=\textwidth]{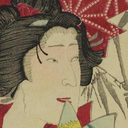}
\end{subfigure}\hfill
\begin{subfigure}[t]{0.715\textwidth}
    \includegraphics[width=\textwidth]{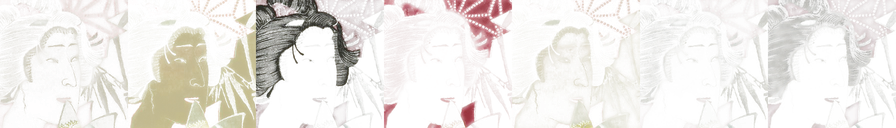}
\end{subfigure}\hfill
\begin{subfigure}[t]{0.102\textwidth}
    \includegraphics[width=\textwidth]{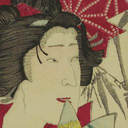}
\end{subfigure}\hfill

\begin{subfigure}[t]{0.102\textwidth}
    \includegraphics[width=\textwidth]{./images/null.png}
    \subcaption*{\small Input}
\end{subfigure}\hfill
\begin{subfigure}[t]{0.715\textwidth}
    \includegraphics[trim=0 0 0 30px, clip, width=\textwidth]{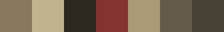}
    \subcaption*{\small Decomposed Layers and Color Palette}
\end{subfigure}\hfill
\begin{subfigure}[t]{0.102\textwidth}
    \includegraphics[width=\textwidth]{./images/null.png}
    \subcaption*{\small Reassembled}
\end{subfigure}\hfill

\ks{}

\caption{Soft color separation takes as input \ukiyoe{} paintings (left) and a color palette (middle), and produces decomposed layers of homogeneous colors (middle). These layers can be used as the inferred woodblocks for corresponding colors and composed back to a reassembled painting (right) resembling the original one.
We infer the color palette by applying K-means clusttering~\protect\cite{lloyd1982least} on the input painting's pixels. 
}

\label{fig:color.separation}

\ks{}

\end{figure*}
\begin{figure}[ht]
\centering

\ks{}

\newcommand{\barratio}{0.042}
\newcommand{\subratio}{0.207}

\hfill
\begin{subfigure}[t]{\subratio\columnwidth}
    \includegraphics[width=\textwidth]{./images/null.png}
\end{subfigure}\hfill
\begin{subfigure}[t]{\barratio\columnwidth}
    \includegraphics[width=\textwidth]{./images/null.png}
\end{subfigure}\hfill
\begin{subfigure}[t]{\subratio\columnwidth}
    \includegraphics[width=\textwidth]{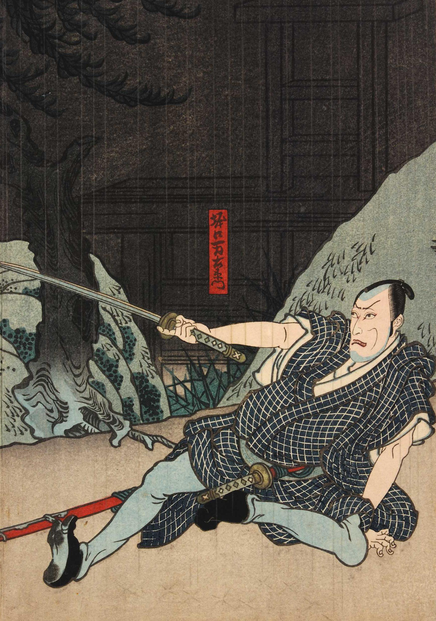}
\end{subfigure}\hfill
\begin{subfigure}[t]{\subratio\columnwidth}
    \includegraphics[width=\textwidth]{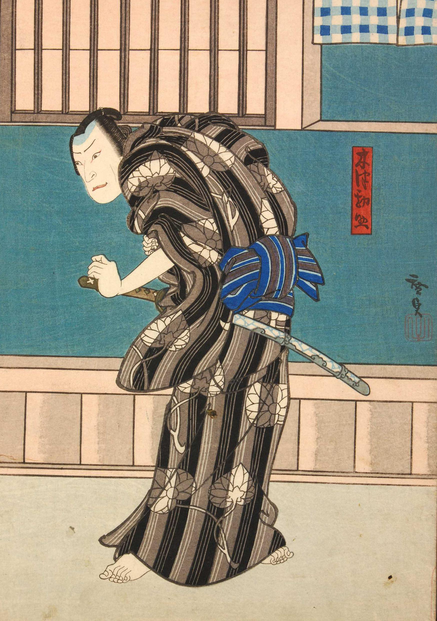}
\end{subfigure}\hfill
\begin{subfigure}[t]{\subratio\columnwidth}
    \includegraphics[width=\textwidth]{./images/null.png}
\end{subfigure}\hfill

\hfill
\begin{subfigure}[t]{\subratio\columnwidth}
    \includegraphics[width=\textwidth]{./images/null.png}
\end{subfigure}\hfill
\begin{subfigure}[t]{\barratio\columnwidth}
    \includegraphics[width=\textwidth]{./images/null.png}
\end{subfigure}\hfill
\begin{subfigure}[t]{\subratio\columnwidth}
    \includegraphics[width=\textwidth]{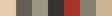}
    \includegraphics[width=\textwidth]{./images/null.png}
\end{subfigure}\hfill
\begin{subfigure}[t]{\subratio\columnwidth}
    \includegraphics[width=\textwidth]{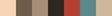}
\end{subfigure}\hfill
\begin{subfigure}[t]{\subratio\columnwidth}
    \includegraphics[width=\textwidth]{./images/null.png}
\end{subfigure}\hfill

\vspace*{-5px}

\hfill
\begin{subfigure}[t]{\subratio\columnwidth}
    \includegraphics[width=\textwidth]{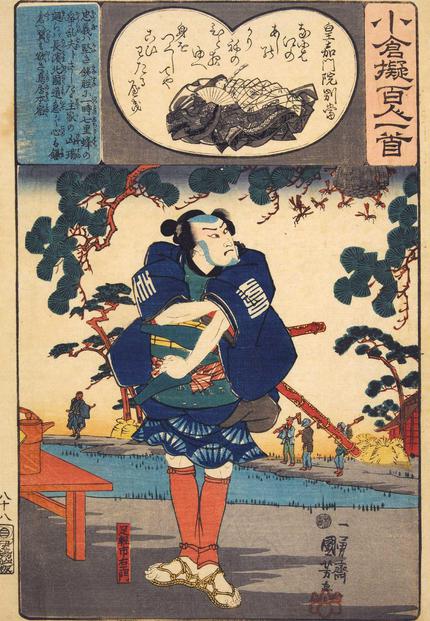}
    \caption{}
\end{subfigure}\hfill
\begin{subfigure}[t]{\barratio\columnwidth}
    \includegraphics[width=\textwidth]{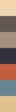}
    \caption{}
\end{subfigure}\hfill
\begin{subfigure}[t]{\subratio\columnwidth}
    \includegraphics[width=\textwidth]{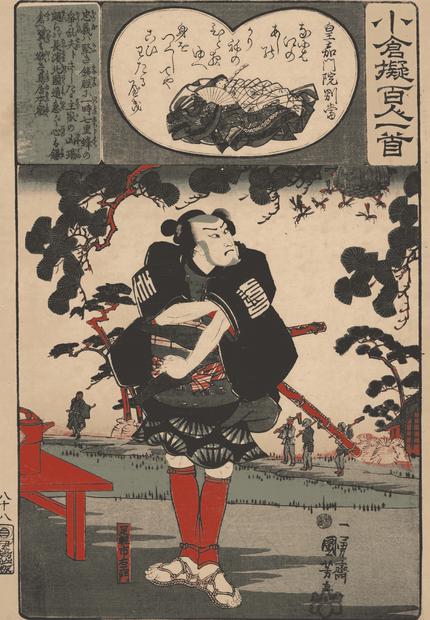}
    \caption{}
\end{subfigure}\hfill
\begin{subfigure}[t]{\subratio\columnwidth}
    \includegraphics[width=\textwidth]{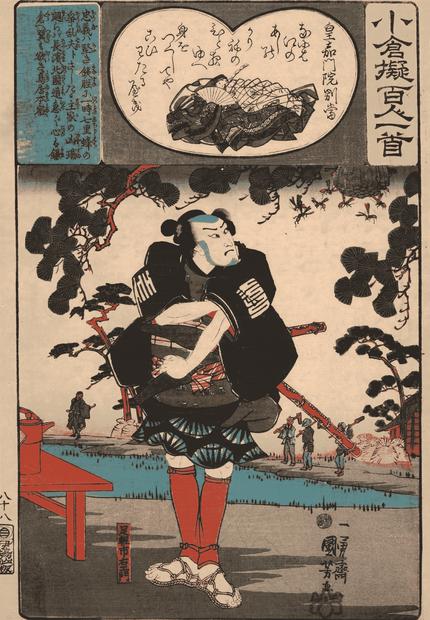}
    \caption{}
\end{subfigure}\hfill
\begin{subfigure}[t]{\subratio\columnwidth}
    \includegraphics[width=\textwidth]{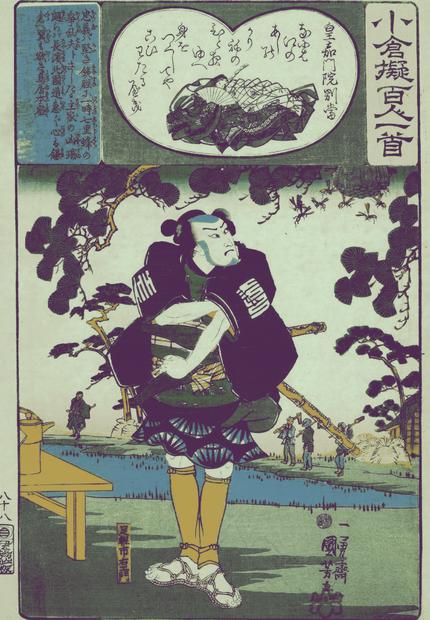}
    \caption{}
\end{subfigure}\hfill

\caption{Decomposing an \ukiyoe{} painting \textbf{(a)} with color palette \textbf{(b)} and recoloring, which could be done automatically \textbf{(c, d)} using color palettes inferred from the reference images, or manuallly with Adobe After Effects \textbf{(e)}.}

\label{fig:color.recoloring}

\ks{}

\end{figure}

\subsubsection{Comparing \ukiyoe{}, Ehon and Human Faces}

As exemplary paintings in Figure~\ref{fig:year.distribution} (c) show, \ukiyoe{} paintings are characterized by their particular facial geometry, which could potentially be different from other art genres or photo-realistic human faces.
To quantify such observation, we conduct unsupervised (PCA, T-SNE, UMAP) clustering of \ukiyoe{} (popular in the 19\superth{} century) faces, Ehon / Emakimono (another Japanese artworks genre popular in the 16\superth{} to 17\superth{} century) faces, and realistic human faces.

Concretely, we use \Kaokore{}~\cite{tian2020kaokore} for Ehon / Emakimono faces, as well as human face photos collected in FFHQ~\cite{karras2019style} dataset that are published under \href{https://creativecommons.org/licenses/by/2.0/}{CC BY 2.0} license.
In Figure~\ref{fig:ukiyoe_vs_kaokore_vs_human.quant}, we can observe that the geometry of \ukiyoe{} faces is different from \Kaokore{}, and only share similarities to a small section of realistic human faces.
This observation confirms the uniqueness of Japanese artworks' way of portraying humans compared to the real-world image and shows that the development of Japanese artworks over time is a drastic one. 

\subsection{Study Ukiyo-e \emph{Style} through Color Separation}

\ukiyoe{} printings distinguish themselves from other traditional Japanese artworks by the very manner of producing.
Unlike Ehon, which is targeted at a small audience and thus painted by hand, \ukiyoe{} is mass-produced using woodblock printing after the painter finishes the master version.
As shown in a modern reproducing process 
(\href{https://www.adachi-hanga.com/ukiyo-e-en/quality/flow/index_en.html}{\color{link} link}),
multiple woodblocks are carved, each for a portion in the image with a single color, and are printed sequentially with corresponding inks onto the final canvas. 
Unfortunately, such a process for a given \ukiyoe{} painting is not precisely reproducible since the underlying woodblocks are vulnerable, easily worn-out, and often discarded after a certain number of prints.
Thus from an art research point of view, it would be interesting to recover the above-mentioned separated portions for a given \ukiyoe{} painting with only access to the image itself.

We address this challenge by framing it as a \emph{soft color segmentation}~\cite{aksoy2017unmixing} task, which decomposes an input image into several RGBA layers of homogeneous colors.
The alpha channel (``A'' in ``RGBA'') in each layer allows pixels to potentially belong to multiple layers,  which captures ambiguity unavoidable due to imperfect woodblock carving and alignment in multi-pass printing.
In detail, we use state-of-the-art Fast Soft Color Separation (FSCS)~\cite{akimoto2020fast} for efficient processing.
As shown in Figure~\ref{fig:color.separation}, FSCS decomposes \ukiyoe{} paintings into layers of homogeneous colors using color palette. The inferred layers could be interpreted as woodblocks with corresponding colors that could be used for making a particular artwork. 

The decomposition of a painting into multiple layers of homogeneous colors allows us to explore further creativity. 
One example in this direction is recoloring, where we pick a new color for each of the individual layers and compose them into a recolored painting. 
As shown in Figure~\ref{fig:color.recoloring}, the recoloring could be done either automatically using the inferred color palette from other artworks or manually in Adobe After Effects with \emph{alpha add} mode for blending.
The recoloring here serves as an example to study artworks and opens the door to reinterpret them in a new way.

\subsection{Study Jointly Ukiyo-e \emph{Object} and \emph{Style} by Composing Sketch and Color}

As we deals with a dataset focusing on artworks, 
it becomes natural to ask whether we could engage them with approaches invoking creativity and artistic expression.
\begin{figure}[!h]

\centering

\ks{}

\begin{subfigure}[t]{\columnwidth}
    \includegraphics[width=\textwidth]{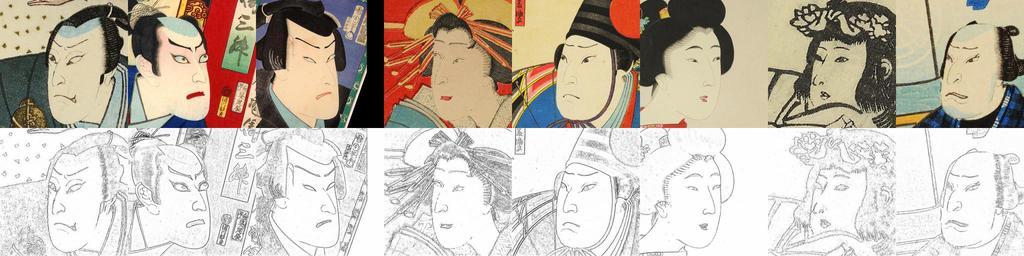}
\end{subfigure}

\ks{}

\caption{Pairs of original \ukiyoe{} faces on the upper row and corresponding line art sketches on the lower row.}
\label{fig:colorization.face.pair.example}

\ks{}

\end{figure}\unskip
\begin{figure}[!h]
    
\centering


\newcommand{\subwidth}{0.24\columnwidth}
\begin{subfigure}[t]{\subwidth{}}
    \includegraphics[width=\textwidth]{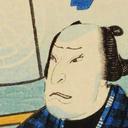}
    \caption{\small Origin}
\end{subfigure}\hfill
\begin{subfigure}[t]{\subwidth{}}
    \includegraphics[width=\textwidth]{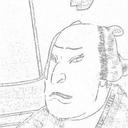}
    \caption{\small Sketch}
\end{subfigure}\hfill
\begin{subfigure}[t]{\subwidth{}}
    \includegraphics[width=\textwidth]{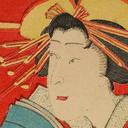}
    \caption{\small Reference}
\end{subfigure}\hfill
    \begin{subfigure}[t]{\subwidth{}}
    \includegraphics[width=\textwidth]{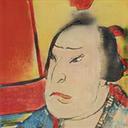}
    \caption{\small Colorized}
\end{subfigure}\hfill

\ks{}

\caption{Colorization on \ukiyoe{} faces.
For a face painting \textbf{(a)}, we extract its line art sketch \textbf{(b)}.
A colorization model takes both the sketch and a reference painting \textbf{(c)}, and produces a colorized painting \textbf{(d)} reflecting the sketch's geometry and the reference's style in colors and textures.}
\label{fig:colorization.face.blend.example}

\ks{}

\end{figure}\unskip
\begin{figure*}[t!]
\centering

\ks{}


\begin{subfigure}[t]{0.513\textwidth}
    \includegraphics[width=\textwidth]{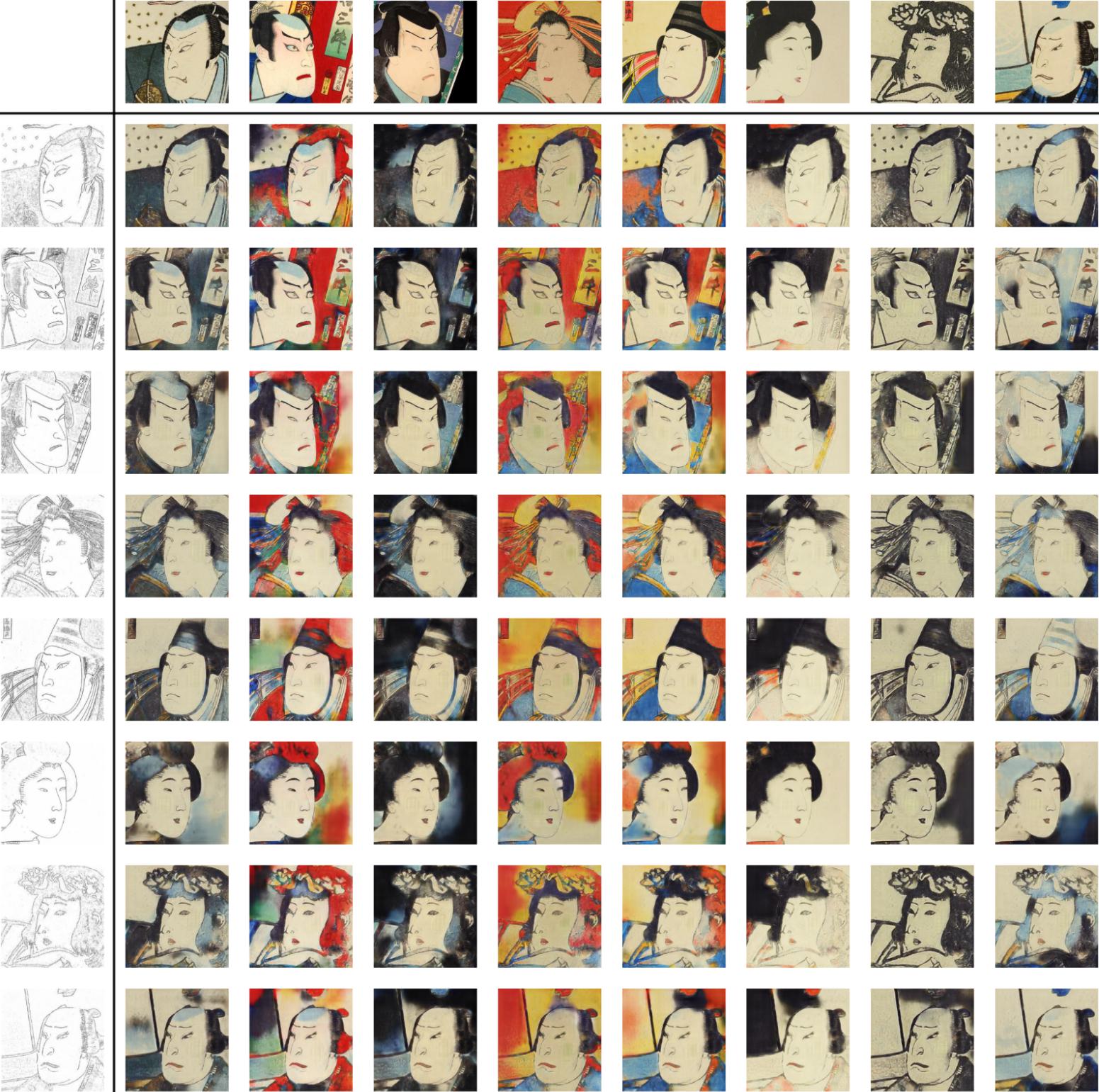}
    \caption{\small \ukiyoe{} faces.}
\end{subfigure}\hfill
\begin{subfigure}[t]{0.387\textwidth}
    \includegraphics[width=\textwidth]{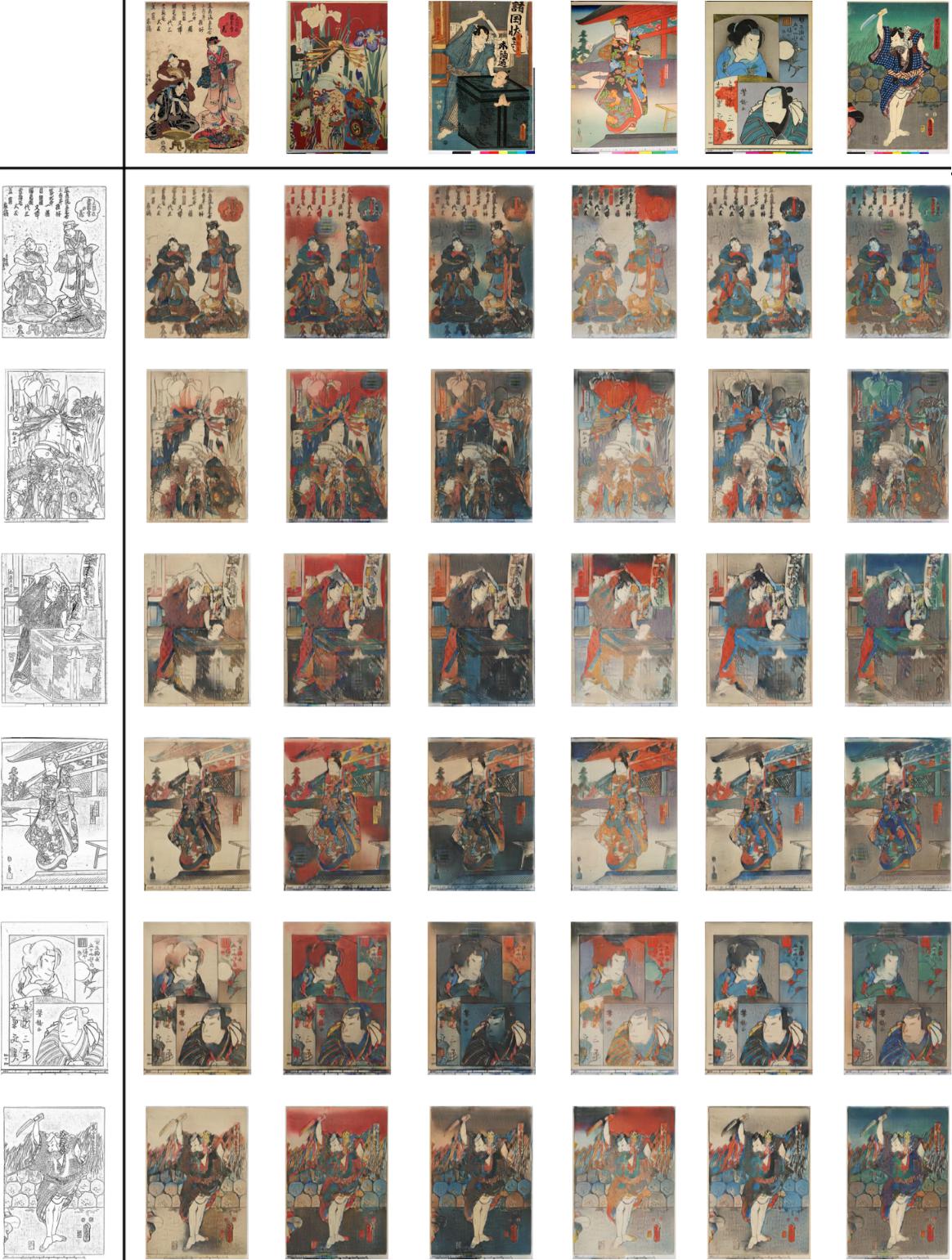}
    \caption{\small Whole \ukiyoe{} paintings.}
\end{subfigure}

\ks{}

\caption{Matrices of blending line art sketches and painting style for \ukiyoe{} faces \textbf{(a)} and whole \ukiyoe{} paintings \textbf{(b)}.
Within a single matrix, each row represents an art line sketch, each column represents the reference image for style, and images at an intersection are the blending results of the corresponding row and column.}

\label{fig:colorization.crossblend}

\ks{}

\end{figure*}\unskip
One direction is to examine whether the recent advances of machine learning models could create structurally sound, or even artistically impressive, results.
In this direction, generative models has been proposed to generate faces in Japanese painting style~\cite{tian2020kaokore} and blend generative models trained on data of different domains by swapping layers of two image generation neural networks~\cite{pinkney2020resolution}.
However, the former lacks controllability in the generation as it can only produce images as a whole, and the latter focuses on transferring across separated, different domains by the nature of its design.

Thus we identify an unbridged gap in the \emph{in-domain} separation of artistically essential aspects. 
In detail, we ask the following question: what is the (dis)entanglement between the \emph{object} and \emph{style} within the \ukiyoe{}. 
Answering this question reveals the relation between \ukiyoe{}'s \emph{object} and \emph{style}. 
Furthermore, it also allows editing one of them while keeping another intact for creative expression.
One way to separate the \emph{object} and \emph{style} is to represent the former with line art sketches for what person/scene is depicted, and the latter with color and texture information showing the painting style. 
They could be composed with a colorization process, 
which blends a sketch as an \emph{object} reference and an image as a reference for instance-level painting \emph{style}.

\begin{figure}[h!]
    
\centering

\ks{}

\begin{subfigure}[t]{\columnwidth}
    \includegraphics[width=\textwidth]{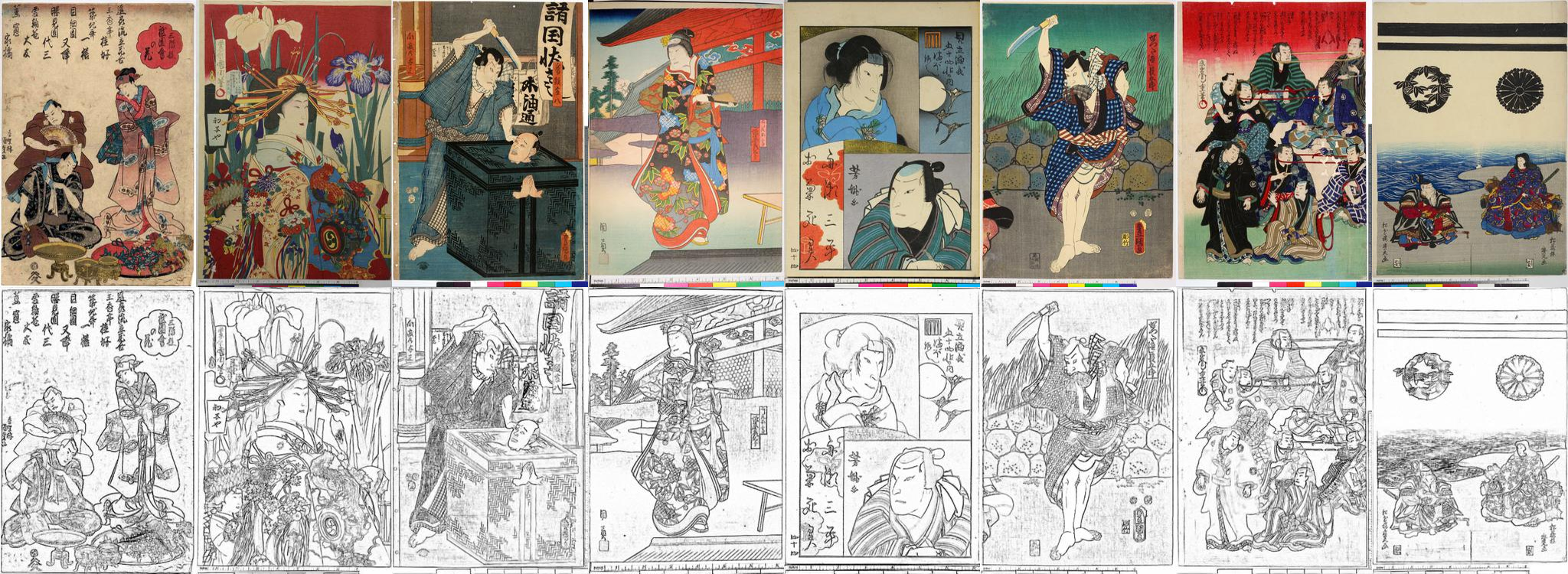}
\end{subfigure}

\vspace*{0.6cm}    

\newcommand{\subwidth}{0.24\columnwidth}
\begin{subfigure}[t]{\subwidth{}}
    \includegraphics[width=\textwidth]{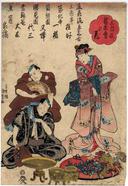}
    \caption{\small Origin}
\end{subfigure}\hfill
\begin{subfigure}[t]{\subwidth{}}
    \includegraphics[width=\textwidth]{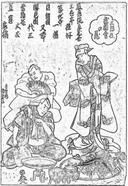}
    \caption{\small Sketch}
\end{subfigure}\hfill
\begin{subfigure}[t]{\subwidth{}}
    \includegraphics[width=\textwidth]{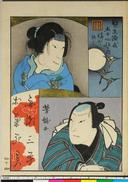}
    \caption{\small Reference}
\end{subfigure}\hfill
    \begin{subfigure}[t]{\subwidth{}}
    \includegraphics[width=\textwidth]{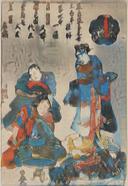}
    \caption{\small Colorized}
\end{subfigure}\hfill

\ks{}

\caption{Top: Pairs of whole \ukiyoe{} painting and line art sketches in the same format as Figure~\ref{fig:colorization.face.pair.example}.
Bottom: Colorization on \ukiyoe{} faces in the same format as Figure~\ref{fig:colorization.face.blend.example}.}

\label{fig:colorization.whole.pair-and-blend.example}

\ks{}

\end{figure}\unskip

\textbf{Face Images}. We extract line art sketches from \ukiyoe{} images using SketchKeras~\cite{zhang2017}, as illustrated in Figure~\ref{fig:colorization.face.pair.example}.
We further train image colorization~\cite{lee2020} using a public-available implementation~\cite{hasegawa2020}.
The whole pipeline is illustrated in Figure~\ref{fig:colorization.face.blend.example}. 
As shown in Figure~\ref{fig:colorization.crossblend}, Since the model learns to separate the \emph{object}, indicated in the sketch image, and the \emph{style}, indicated by reference image, as two orthogonal and composable semantics,
it could blend arbitrary combination of sketch and reference style images.
Such separation could enable future works to help with humanities research on combinations of \ukiyoe{} color and subject. 
For example, in \ukiyoe{} depicting Kabuki, the attributes and colors of the characters are somewhat correlated semantically. 
\begin{figure}[h!]
    
\centering

\ks{}

\includegraphics[width=\columnwidth]{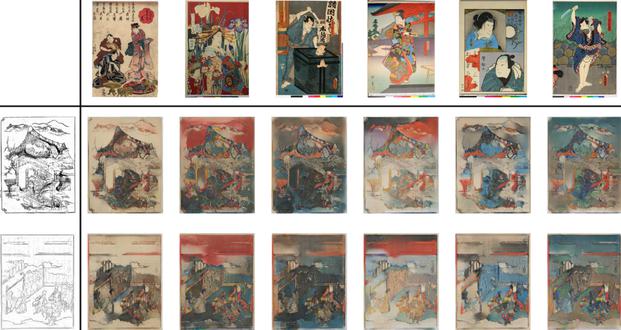}

\ks{}

\caption{Colorization in-the-wild woodblock printing using the model trained on the whole \ukiyoe{} paintings. Each row represents a woodblock printing work, and each column represents the reference image for style.}

\label{fig:coloriztaion.inthewild.woodblockpainting}

\ks{}

\end{figure}\unskip
\begin{figure}[h!]
    
\centering

\ks{}

\newcommand{\subwidth}{1.0\columnwidth}
\begin{subfigure}[t]{\subwidth{}}
    \includegraphics[width=\textwidth]{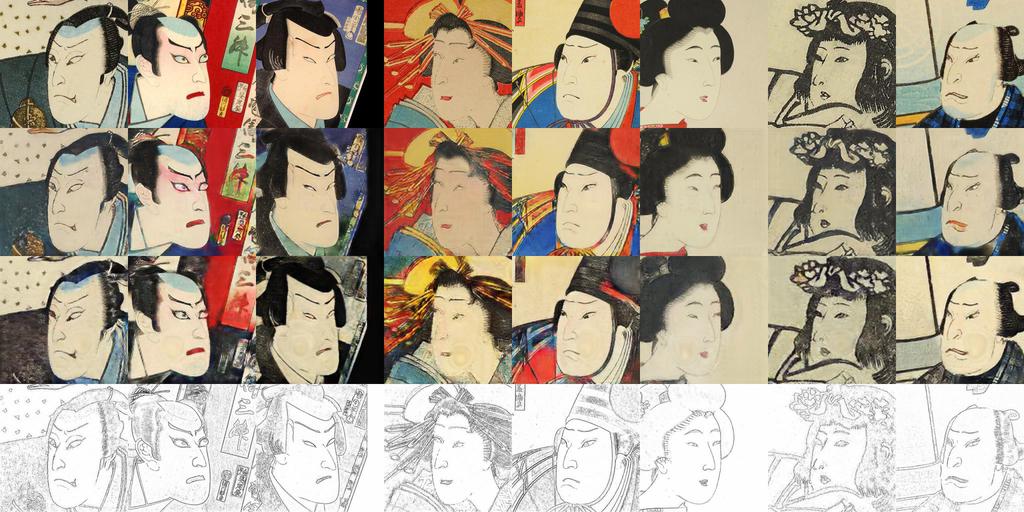}
\end{subfigure}\hfill

\ks{}

\caption{Comparation of conditional and the unconditional colorization method. The former uses style reference images while the latter does not.
Four rows are the ground truth color image, conditional colorization, unconditional colorization, and line art sketches, respectively.
}
\label{fig:colorization.ref_vs_nonref}
    
\ks{}

\end{figure}\unskip
Therefore swapping colors can change the meaning of scenes and people in the painting.
We envision that discoveries could be made by studying how the impression of \ukiyoe{} paintings changes through the process of swapping colors.

\textbf{Whole Painting}. We go beyond faces and work on whole \ukiyoe{} painting images.
By employing the same pipeline to the whole painting images, as shown in Figure~\ref{fig:colorization.whole.pair-and-blend.example}, the model can be further leveraged to colorize in-the-wild woodblock printing images, as Figure~\ref{fig:coloriztaion.inthewild.woodblockpainting} shows.
However, while the resulting colorized images are reasonable, they are of lower quality than those of faces.
Such observation is anticipated since the whole \ukiyoe{} painting
is more complex than face in many ways, like
topics and topological configuration of objects, which presents a much more challenging task for colorization.
This issue could be further exaggerated by the discrepancy between the \ukiyoe{} domain where the model is trained and the woodblock painting domain where the model is applied.
We would leave higher quality, whole \ukiyoe{} painting colorization for future study.

\textbf{Conditional vs. Unconditional Colorization}.
While we choose to use a conditional colorization method, which produces results from a sketch \emph{and} a reference image for color and styles,
it is also worth considering a simper, unconditional colorization method that directly generates the results from a sketch, such as \pixtopixhd{}~\cite{wang2018pix2pixHD}.
This alternation, however, suffers from the inability to control the color and style of the generated image.
Moreover, as we show in Figure~\ref{fig:colorization.ref_vs_nonref}, the unconditional colorization method produces worse colorization results than the conditional colorization method~\cite{lee2020}.
We argue that this is expected since the former method has to fall back to safe colors that valid for any \ukiyoe{} images,
while the latter could make a wiser choice based on the reference images.

\subsubsection{Discussion}
We show that \ukiyoe{} paintings can be studied by (1) representing \emph{object} with line art sketches, (2) representing \emph{style} as a color reference image, and (3) composing them using colorization.
This pipeline provides a clear separation of two semantics important in the art research and allows further creativity through compositions of both in unseen ways.
As it is just one possible way of studying the interaction between the \emph{object} and the \emph{style}, we expect further works could explore different forms of creative expression.

\section{Conclusion}

In this work, we propose to bridge the machine learning and humanities research on the subject of \ukiyoe{} paintings.
Besides the presented dataset with coherent labels and annotations, 
we also show their value in the quantification approach to humanities research, 
Furthermore, we demonstrate that machine learning models in a creative setting could address art-style research problems.
\section{Acknowledgement}

We thank Hanjun Dai, David Ha, Yujing Tang, Neil Houlsby, and Huachun Zhu for their comments and helpful suggestions.

\bibliographystyle{iccc}
\bibliography{references}

\end{CJK}
\end{document}